\documentclass{article} 
\usepackage{flashomni_2026_conference,times}

\usepackage[utf8]{inputenc}
\usepackage[T1]{fontenc}

\usepackage{color,xcolor}
\usepackage{epsfig}
\usepackage{graphicx}

\usepackage{adjustbox}
\usepackage{array}
\usepackage{booktabs}
\usepackage{colortbl}
\usepackage{float,wrapfig}
\usepackage{hhline}
\usepackage{multirow}
\usepackage{subcaption}
\usepackage[font=small]{caption}
\usepackage{makecell}

\usepackage{amsmath,amsfonts,amsthm,amssymb}
\theoremstyle{definition}
\newtheorem{definition}{Definition}

\usepackage{bm}
\usepackage{nicefrac}
\usepackage{microtype}

\usepackage{changepage}
\usepackage{extramarks}
\usepackage{fancyhdr}
\usepackage{lastpage}
\usepackage{setspace}
\usepackage{soul}
\usepackage{xspace}
\usepackage{indentfirst}
\usepackage{pifont}
\usepackage{cuted}
\usepackage{wrapfig}
\definecolor{cvprblue}{rgb}{0.21,0.49,0.74}

\usepackage[pagebackref,breaklinks,colorlinks,citecolor=cvprblue]{hyperref}
\usepackage{url}

\usepackage[linesnumbered,ruled,vlined]{algorithm2e}
\usepackage{enumitem}

\usepackage{wasysym}
\usepackage{duckuments}
\usepackage{pifont}
\usepackage{duckuments}
\usepackage{fancyvrb}
\usepackage{fvextra}

\usepackage{bm}
\usepackage{listings}
\usepackage[normalem]{ulem}
\usepackage{amsmath,amsfonts,bm}

\def\eqref#1{equation~\ref{#1}}

\def\1{\bm{1}}

\DeclareMathAlphabet{\mathsfit}{\encodingdefault}{\sfdefault}{m}{sl}
\SetMathAlphabet{\mathsfit}{bold}{\encodingdefault}{\sfdefault}{bx}{n}

\newcolumntype{L}[1]{>{\raggedright\let\newline\\\arraybackslash\hspace{0pt}}m{#1}}
\newcolumntype{C}[1]{>{\centering\let\newline\\\arraybackslash\hspace{0pt}}m{#1}}
\newcolumntype{R}[1]{>{\raggedleft\let\newline\\\arraybackslash\hspace{0pt}}m{#1}}

\newcommand{\ignorethis}[1]{}

\definecolor{citecolor}{rgb}{34,139,34}
\definecolor{mydarkblue}{rgb}{0,0.08,1}
\definecolor{mydarkgreen}{rgb}{0.02,0.6,0.02}
\definecolor{mydarkred}{rgb}{0.8,0.02,0.02}
\definecolor{mydarkorange}{rgb}{0.40,0.2,0.02}
\definecolor{mypurple}{RGB}{111,0,255}
\definecolor{myred}{rgb}{1.0,0.0,0.0}
\definecolor{mygold}{rgb}{0.75,0.6,0.12}
\definecolor{mydarkgray}{rgb}{0.66,0.66,0.66}

\newcommand{\jt}[1]{\textcolor{blue}{{#1}}\xspace}

\definecolor{comment_color_2}{RGB}{64,128,128}
\newcommand{\LineComment}[1]{\vspace*{0.1em}\textcolor{comment_color_2}{\textit{\# #1}}}

\def\method{FlashOmni\xspace}

\iclrfinalcopy

\title{\method: A Unified Sparse Attention Engine for Diffusion Transformers}

\author{\textbf{Liang Qiao}$^{1}$\thanks{Equal contribution. \texttt{\{ql1an9, dy329261472\}}@mail.ustc.edu.cn} 
\quad
\textbf{Yue Dai}$^{1}$\footnotemark[1]
\quad
\textbf{Yeqi Huang}$^{2}$
\quad
\textbf{Hongyu Kan}$^{3}$
\quad
\textbf{Jun Shi}$^{1}$
\quad
\textbf{Hong An}$^{1}$ \\
$^1$University of Science and Technology of China
\quad
$^2$University of Edinburgh \\
$^3$University of Virginia \\
\textbf{Code}: \url{https://github.com/qiaolian9/FlashOmni}
}

\begin{document}

\maketitle

\begin{abstract}

Multi-Modal Diffusion Transformers (DiTs) demonstrate exceptional capabilities in visual synthesis, yet their deployment remains constrained by substantial computational demands. To alleviate this bottleneck, many sparsity‑based acceleration methods have been proposed. However, their diverse sparsity patterns often require customized kernels for high-performance inference, limiting universality. We propose \textbf{FlashOmni}, a unified sparse attention engine compatible with arbitrary DiT architectures. FlashOmni introduces flexible \emph{sparse symbols} to standardize the representation of a wide range of sparsity strategies, such as feature caching and block‑sparse skipping. This unified abstraction enables the execution of diverse sparse computations within a single \textit{attention kernel}.  In addition, FlashOmni designs optimized \textit{sparse GEMMs} for attention blocks, leveraging sparse symbols to eliminate redundant computations and further improve efficiency.  Experiments demonstrate that FlashOmni delivers near‑linear, closely matching the sparsity ratio speedup  (1:1) in attention and GEMM‑$Q$, and achieves \(2.5\times\)–\(3.8\times\) acceleration in GEMM‑$O$ (max peaking at about 87.5\% of the theoretical limit). Applied with a multi‑granularity sparsity strategy, it enables the Hunyuan model (33K) to achieve about \(1.5\times\) end‑to‑end acceleration without degrading visual quality.

\end{abstract}
\section{Introduction}
\looseness=-1
Diffusion Transformers (DiTs)~\citep{peebles2023scalable} have achieved remarkable progress in high-fidelity visual generation~\citep{esser2024scaling, flux1,kong2024hunyuanvideo, li2024hunyuan,batifol2025flux} by leveraging attention. However, its high computational complexity constrains inference efficiency, a challenge that grows with model scale, especially for high-resolution image and long video generation scenarios. To address this, various acceleration approaches have been developed, with sparsity‑based methods standing out as the most universally applied techniques for their rapid, training‑free deployment.

There are two distinct categories of sparse acceleration methods based on the granularity of sparsity applied during attention computation. \textit{(i) Feature caching}~\citep{chen2024delta, selvaraju2024fora, zou2024accelerating, zouaccelerating,liu2025reusing} primarily exploits the feature similarity between adjacent timesteps, focusing on reusing or predicting the whole computations of selected tokens across steps to reduce the per‑step computational burden by caching the corresponding feature.  \textit{(ii) Block‑sparse skipping} leverages the inherent sparsity of attention computation, where many softmax operation outputs approach zero~\citep{deng2024attention}, to skip unimportant block‑tile computations along the reduction axis via various evaluation strategies.  As attention patterns vary considerably across different tasks, multiple sparse attentions have emerged, such as dynamic sparse attention~\citep{xisparse, zhangspargeattention, xia2025training, xu2025xattention, yang2025sparse} and pattern-based sparse attention~\citep{yuan2024ditfastattn, zhang2025ditfastattnv2}, each providing a task-specific sparsity design.

\begin{figure}[t]
    \centering
    \includegraphics[width=\linewidth]{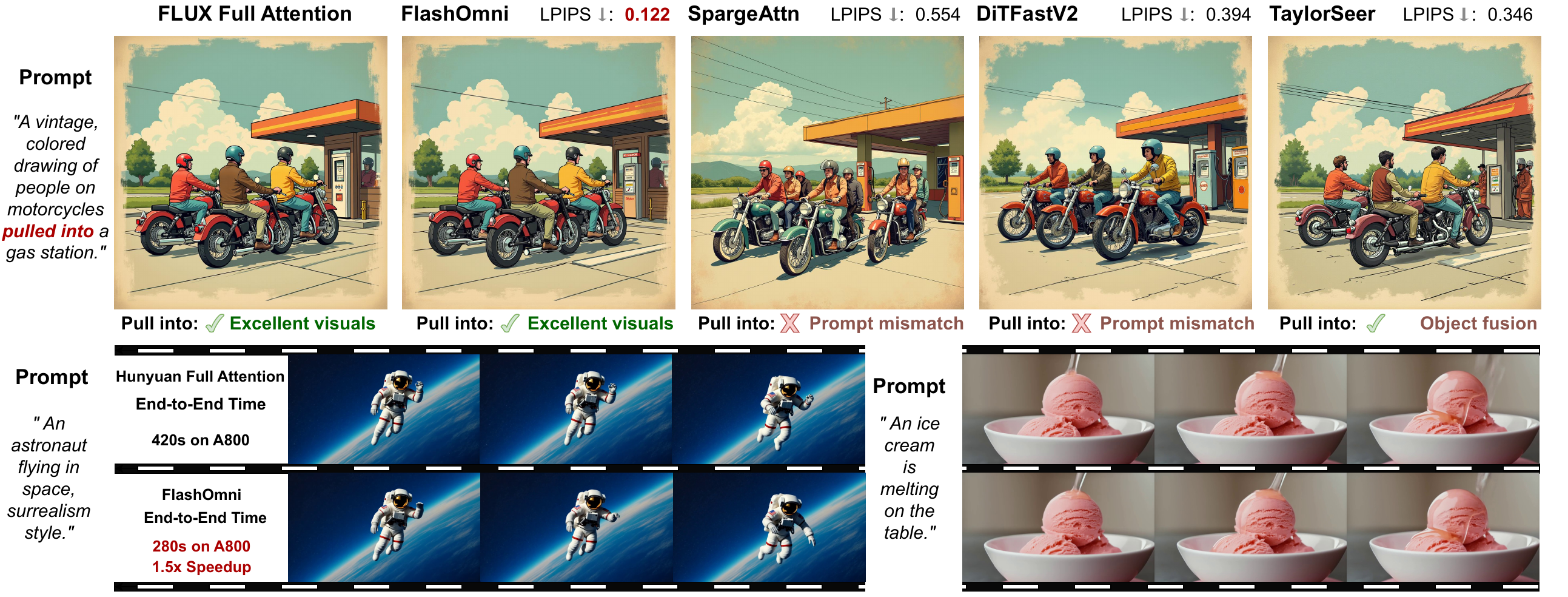}
    \vspace{-15pt}
	\caption{Visualization of different acceleration methods on FLUX and FlashOmni's speedup on HunyuanVideo.
    }
    \vspace{-10pt}
    \label{fig:FlashOmni_vision_demo}
\end{figure}
\looseness=-1
Despite promising progress, sparse acceleration for DiTs still faces several limitations:
\textit{(i) Inconsistent sparsity granularity.} Existing methods vary from coarse-grained caching to fine-grained block skipping. Applying a uniform strategy across all tokens often harms quality, while combining strategies remains difficult without a unified framework.
\textit{(ii) Fragmented design space.} Current methods introduce sparsity patterns tailored to specific tasks (e.g., dynamic vs. pattern-based attention). This fragmentation complicates the search for optimal strategies and prevents reuse across different applications.
\textit{(iii) Lack of kernel generality.} To achieve efficiency, most sparse approaches require dedicated kernels optimized for specific sparsity structures. This reduces flexibility, increases engineering overhead, and hinders scalable deployment.

\looseness=-1
To address these challenges, we propose \method, a unified sparse attention engine for diffusion transformers. \method abstracts the multi‑step denoising using multi‑granularity sparsity as an \textit{"Update‑Dispatch"} paradigm and introduces several key designs: \textit{(i) Unified sparse symbols.} We introduce compact 8-bit sparse symbols to represent multiple levels of sparsity in a unified format. These symbols guide the selective update of cached features, enabling flexible multi-granularity integration. \textit{(ii) General sparse attention kernel.} We design a single kernel that decodes sparse symbols at runtime and efficiently executes diverse sparsity strategies. This eliminates fragmentation by supporting arbitrary sparsity patterns within one engine. \textit{(iii) Optimized sparse GEMMs.} We further develop GEMM-$Q$ and GEMM-$O$, which leverage sparse symbols to eliminate redundant computations along spatial and reduction dimensions, while also improving feature-cache storage logic.

\looseness=-1
Our experiments demonstrate that our sparse kernel design achieves near-linear speedup with increasing sparsity, reaching a one-to-one acceleration with theoretical computation reduction in GEMM-$Q$ and attention, up $2.5\times$ to $3.8\times$ acceleration in GEMM-$O$ (max peaking at about 87.5\% of the theoretical limit), and delivering $2\times$ attention speedup with nearly $1.5\times$ end-to-end gain in Hunyuan built upon \method.

Our contributions can be summarized as follows:
\begin{itemize}[leftmargin=*]
\vspace{-5pt}
\looseness=-1
\item We propose \method, a unified sparse attention engine for DiTs. \method enables highly efficient inference with multi‑granularity sparsity by abstracting an \textit{"Update‑Dispatch"} paradigm.

\looseness=-1
\item \method unifies multi‑granularity sparse strategies with flexible sparse symbols and provides a general attention kernel for supporting efficient arbitrary sparse computation. 
\looseness=-1
\item \method designs sparse GEMMs for the attention module’s linear layer to eliminate redundant computations caused by feature caching and to improve cache storage logic.

\end{itemize}
\section{Related Works}

\begin{wrapfigure}{r}{0.25\linewidth}
    \vspace{-35pt}
    \centering
    \includegraphics[width=\linewidth]{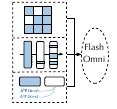}
    \vspace{-20pt}
    \caption{\looseness=-1 Typical sparse methods for DiTs.}
    \vspace{-15pt}
    \label{fig:relation}
\end{wrapfigure}
\looseness=-1
\textbf{Transformers in Diffusion Models.}  Diffusion Transformers~\citep{peebles2023scalable}, which follow scaling laws, have achieved notable success in image~\citep{chen2024delta, li2024hunyuan} and video generation~\citep{zheng2024open,hacohen2024ltx}. Recently, the Multimodal Diffusion Transformer (MMDiT), introduced by SD3~\citep{esser2024scaling}, offers greater advantages. It replaces the cross‑attention module of DiTs by independently projecting visual inputs and text embeddings before concatenating them for self‑attention. Prominent adopters include SD3 and Flux series~\citep{flux1, batifol2025flux} in image synthesis, and CogVideoX~\citep{yang2024cogvideox}, HunyuanVideo~\citep{sun2024hunyuan}, and Mochi‑1~\citep{genmo2024mochi} in video generation. However, iterative sampling causes significant overhead in large and complex DiTs, limiting real‑time applicability. Two primary sparse acceleration strategies are detailed below and illustrated in Figure~\ref{fig:relation}.

\looseness=-1
\textbf{Feature Caching.}
Feature caching leverages representation similarity between adjacent timesteps in DiTs and can be categorized into layer‑wise and token‑wise strategies. approaches~\citep{ma2024deepcache,so2312frdiff,li2023faster,wimbauer2024cache} mitigate redundant computation via caching, interpolation, or low‑frequency updates, with extensions that cache activations in both attention and MLP layers~\citep{chen2024delta,selvaraju2024fora}. TeaCache~\citep{liu2025timestep} further refines this by dynamically estimating timestep‑dependent differences, while TaylorSeer~\citep{liu2025reusing} enhances generation quality by forecasting future features from historical ones. In contrast, token‑wise methods target the importance of individual tokens. ToCa~\citep{zouaccelerating} updates cached features based on attention‑derived token scores, whereas Duca~\citep{zou2024accelerating} approximates attention weights using the norm of the value matrix to estimate token importance.

\looseness=-1
\textbf{Block-Sparse Skipping.} Attention computation often exhibits inherent sparsity, with many softmax outputs approaching zero—a property widely exploited in ViTs~\citep{beltagy2020longformer,hassani2023neighborhood,child2019generating} and LLMs~\citep{zhang2023h2o,xiao2023efficient,fu2024moa}. Sparse attention mechanisms for accelerating DiTs can be divided into static and dynamic approaches. Static methods predefine sparse patterns offline, such as prioritizing recent tokens. DiTFastAttn~\citep{yuan2024ditfastattn} combines sliding‑window patterns with attention sharing, while Sparse‑vDiT~\citep{chen2025sparse} classifies attention heads into three predefined patterns. DiTFastAttnV2~\citep{zhang2025ditfastattnv2} employs Arrow Attention with caching to enable head‑wise selection in MMDiT, alleviating window constraints. Dynamic methods instead construct sparsity masks at runtime. Xattention~\citep{xu2025xattention} uses the sum of antidiagonal values in the attention matrix to provide a powerful proxy for block importance. SpargeAttention~\citep{zhangspargeattention} derives masks from QK embeddings without relying on predefined patterns, whereas SVG2~\citep{yang2025sparse} uses semantic‑aware permutation to capture critical tokens better and reduce redundant computation.

\begin{wrapfigure}{r}{0.3\linewidth}
    \vspace{-47pt}
    \centering
    \includegraphics[width=\linewidth]{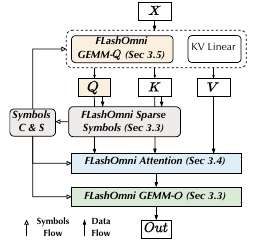}
    \vspace{-20pt}
    \caption{\looseness=-1 \method design.}
    \vspace{-30pt}
    \label{fig:overview1}
\end{wrapfigure}
\section{Method}
\looseness=-1
In this section, we first outline the sparse acceleration methods for text-to-vision diffusion transformers, then introduce the \method framework (Figure~\ref{fig:overview1}), which incorporates novel designs across unified sparse symbols, sparse attention, and sparse GEMMs, followed by detailed descriptions.

\subsection{Preliminary}
\looseness=-1
Text-to-vision diffusion transformers (e.g., MMDiT) experience attention bottlenecks due to concatenating fixed-length text tokens ($N_{\text{text}}$) with resolution-dependent visual tokens ($N_{\text{vision}}$), producing a joint attention map with four regions: text-to-text, $v\!\rightarrow\!t$, $t\!\rightarrow\!v$, and vision-to-vision. Standard attention computes $S=QK^{\top}/\sqrt{d}$, $P=\mathrm{Softmax}(S)$, $O=PV$, with $N=N_{\text{text}}+N_{\text{vision}}$. FlashAttention~\citep{daoflashattention} accelerates this by block-partitioning $Q$, $K$, $V$ (sizes $b_q$, $b_k$) and applying online softmax~\citep{milakov2018online}. Acceleration strategies generally apply one sparsity pattern per block, ranging from coarse- to fine-grained:  

\begin{definition}[Logical Block Sparse Masks]  
$M_c \in \{0,1\}^{\lceil N/b_q \rceil}$ marks cached output blocks; $M_s \in \{0,1\}^{\lceil N/b_q \rceil \times \lceil N/b_k \rceil}$ marks skipped $Q_iK_j^\top$ and $\widetilde{P}_{ij}V_j$ computations.  
\end{definition}  

\begin{definition}[Sparse Strategies]  
Feature caching reuses $O_i$ when $M_c[i]=0$, refreshing masks and cache at \textit{Update} steps. Block-sparse skipping omits pairs with $M_s[i,j]=0$, with dynamic masks updated from the latest $Q$, $K$, and static masks pre-tuned offline for zero update cost.  
\end{definition}

\begin{figure}[t]
    \centering
    \includegraphics[width=\linewidth]{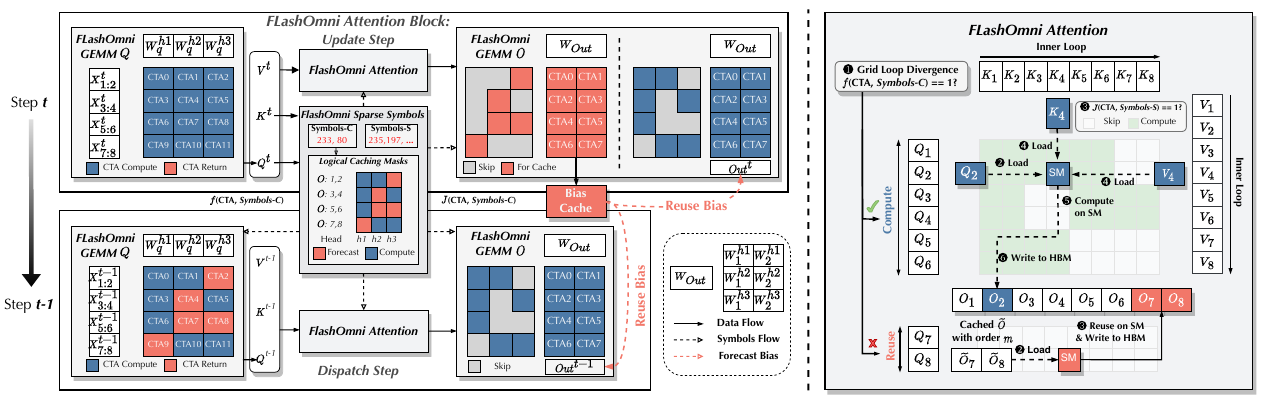}
    \vspace{-15pt}
	\caption{Detailed workflow of the \method framework: incorporating unified sparse symbols and sparse kernels (general sparse attention and GEMMs). Unified sparse symbols are refreshed only at \textit{Update} timesteps, providing sparse guidance for corresponding sparse kernel executions at \textit{Dispatch} timesteps.
    }
    \vspace{-10pt}
    \label{fig:FlashOmni_overview}
\end{figure}
\subsection{\method Overview}
\looseness=-1
Figure~\ref{fig:FlashOmni_overview} presents the operational flow of \method, which adopts an \textit{"Update-Dispatch"} framework to integrate and execute two sparsity strategies across two primary phases:  

\looseness=-1
\textit{Update:}  Given a sequence of adjacent timesteps \(\{t, t-1, \dots, t-\mathcal{N}\}\), \method first refreshes the sparse symbols and feature cache at the current step \(t\). Using the newly computed \(Q\) and \(K\), a tailored sparsity-selection policy determines the sparsity type for each token block in the upcoming steps. Full attention is then applied to update the feature cache. When an output projection is present, \method additionally leverages GEMM-\(O\)  to optimize cache updates further.

\looseness=-1
\textit{Dispatch}: Utilizing the sparse symbols produced in the Update phase, \method accelerates attention computation over the following \(\mathcal{N}\) timesteps \(\{t-1, \dots, t-\mathcal{N}\}\). Within the attention module, Cooperative Thread Arrays (CTA) adopt specialized computation modes, enabling different sparsity granularities for their respective block tiles. Guided by the caching symbols, \method also applies GEMM-\(Q\) and GEMM-\(O\) optimizations, removing redundant operations.

\subsection{\method Sparse Symbols}
\looseness=-1
\method employs two sparse symbols, \(\bm{\mathcal{S}_{c}}\) and \(\bm{\mathcal{S}_{s}}\), serving as a unified representation for multiple sparsity approaches, including feature caching and block-sparse skipping. Using MMDiT as a case study, we examine the joint attention map to determine the applicability of sparsity at two distinct granularities and derive the combination strategy adopted in \method.  

\looseness=-1  
\texttt{Observation 1}. In text-to-vision diffusion transformers, the \mbox{\(v\!\rightarrow\!t\)} and \mbox{\(t\!\rightarrow\!v\)} regions of the attention map are essential for reliable multimodal fusion. At each timestep, \mbox{\(v\!\rightarrow\!t\)} updates text tokens via \mbox{\(P_{v\rightarrow t}V_{\text{vision}}\)}, embedding visual context into \mbox{\(O_{\text{text}}\)}. Conversely, \mbox{\(t\!\rightarrow\!v\)} updates vision tokens via \(P_{t\rightarrow v}V_{\text{text}}\), injecting textual guidance into \(O_{\text{vision}}\). These two interactions are complementary: omitting \(\mbox{\(v\!\rightarrow\!t\)}\) prevents text from perceiving visual changes, whereas omitting \(\mbox{\(t\!\rightarrow\!v\)}\) leads to vision outputs misaligned with textual prompts. Empirically, we observe that caching image tokens that significantly influence text tokens, or are strongly affected by control-signal tokens, degrades cross-modal consistency. Therefore, \method excludes such tokens from caching to ensure timely and accurate multimodal updates.

\begin{wrapfigure}{r}{0.3\linewidth}
    \vspace{-10pt}
    \centering
    \includegraphics[width=\linewidth]{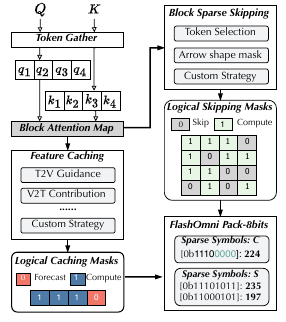}
    \vspace{-20pt}
    \caption{\looseness=-1 Example of \method sparse symbols generation for a single head of attention.}
    \vspace{-15pt}
    \label{fig:FalshOmni_SS}
\end{wrapfigure}
\looseness=-1
\textit{Logical Masks Generation.}  
\method first infers sparsity patterns from the attention structure and encodes them into logical block-sparse masks. These patterns are derived from a compressed attention map, which serves as the basis for constructing the sparse strategy combination used in this work. Specifically, every \(n\) consecutive \(\{Q_i\}\) and \(\{K_j\}\) blocks (e.g., two blocks in Figure~\ref{fig:FlashOmni_overview}) are aggregated into single tokens \(q\) and \(k\) through token‑gathering operations such as mean pooling, with pooling sizes \(b_q\) and \(b_k\). The compressed tokens \(\{q_i\}\) and \(\{k_j\}\) form a reduced attention map \mbox{\(\widetilde{P} = \mathrm{Softmax}(\widetilde{S})\)}, where \mbox{\(\quad \widetilde{S} = \{q_i\} \{k_j^{T}\} \in \mathbb{R}^{\lceil N/(n b_q) \rceil \times \lceil N/(n b_k) \rceil}\)}. From this representation, we define two metrics for guiding feature caching: (i) \underline{Vision‑to‑Text Contribution} \((\mathcal{C}_{i, v\rightarrow t})\): for vision block \(i\), this measures its contribution to text blocks based on the compressed attention map. Lower values indicate less influence on text tokens and are therefore prioritized for caching. Denoting \(\alpha_{i,j}\) as the \((i,j)\) element in \(\widetilde{P}[:n_t, n_t:]\), where \(n_t = \lceil n_t/(n b_q)\rceil\) is the number of compressed text blocks, the metric is computed as \mbox{\(\mathcal{C}_{i, v\rightarrow t} = \sum_{j=1}^{n_t}{\alpha_{j,i}}\)}. (ii) \underline{Text‑to‑Vision Guidance} \((\mathcal{G}_{i, t\rightarrow v})\): this reflects the extent to which vision block \(i\) is influenced by text blocks. Blocks under stronger textual guidance retain attention computation to preserve responsiveness. With \(\beta_{i,j}\) denoting the \((i,j)\) element in \(\mathrm{Softmax}(\widetilde{P}[n_t:, :n_t]^{T})\), the score is given by \mbox{\(\mathcal{G}_{i, t\rightarrow v} = \sum_{j=1}^{n_t}{\beta_{j,i}}\)}. Given \(\mathcal{C}_{i, t\rightarrow v}\) and \(\mathcal{G}_{i, v\rightarrow t}\), \method selects the indices with the lowest combined scores such that their cumulative sums do not exceed thresholds \(\tau_c \cdot \sum_{i} \mathcal{C}_{i, t\rightarrow v}\) and \(\tau_c \cdot \sum_{i} \mathcal{G}_{i, v\rightarrow t}\), as formalized in Equation~\ref{selection_caching}:  

\vspace{-10pt}
\begin{equation} 
\label{selection_caching}
    \{\, i \mid 
\mathrm{CumSum}_{\uparrow}(\mathcal{C}_{i, t\rightarrow v}) \le \tau_c \cdot \sum_j \mathcal{C}_{j, t\rightarrow v} 
\ \wedge\ 
\mathrm{CumSum}_{\uparrow}(\mathcal{G}_{i, v\rightarrow t}) \le \tau_c \cdot \sum_j \mathcal{G}_{j, v\rightarrow t} 
\,\}
\end{equation}  
\vspace{-15pt}

\looseness=-1
Blocks with minimal scores are cached for use in the next step, while others undergo full attention computation. In the logical caching mask \(M_c[i]\), cached positions are assigned 0 and non‑cached positions 1. For cached blocks, \method employs TaylorSeer~\citep{liu2025reusing} to forecast future features via Taylor series expansion using stored features and their derivatives. For block‑sparse skipping, token selection follows the compressed attention map, aligning with the approach of SpargeAttn~\citep{zhangspargeattention}.

\looseness=-1
\textit{Sparse Symbols Compression.}  
To reduce storage overhead, the logical masks are encoded as 8‑bit compressed sparse symbols: \(\bm{\mathcal{S}_{c}}\) for feature caching and \(\bm{\mathcal{S}_{s}}\) for block‑sparse skipping, as shown in Figure~\ref{fig:FalshOmni_SS}. In each \textit{Update} step, the latest \(Q\) and \(K\) are block‑aggregated to form the attention map, from which masks are generated via the two‑granularity sparsity strategy. For example, \(M_{c}[4]=0\) skips the computation of \(O_7\) and \(O_8\) and reusing previous \(O_7\) and \(O_8\) during the "Dispatch" step. With big‑end alignment, zero‑padding yields the binary \texttt{0b11100000}, stored as the uint8 \texttt{224} for \(\mathcal{S}_{c}\); \(\mathcal{S}_{s}\) can similarly be encoded as \texttt{235} and \texttt{197}.

\begin{wrapfigure}{r}{0.4\columnwidth}
\vspace{-15pt}
\begin{minipage}{0.4\textwidth}
\SetAlgoLined\setlength{\algomargin}{0.6em}
\begin{algorithm}[H]
   \fontsize{7.5}{10}\selectfont
    \LinesNumbered
    \SetAlgoNlRelativeSize{-1}
    \caption{\method Attention}
    \label{flashomni_attention}
    
    \KwIn{}
    \hspace{0.1cm} $\{Q_i\}\in \mathbb{R}^{T_q \times b_q \times d}$;$T_q = {N}/{b_q}$ \\ 
    \hspace{0.1cm} $\{K_i\}, \{V_i\}\in \mathbb{R}^{T_{kv} \times b_k \times d}$;$T_{kv} = {N}/{b_k}$ \\
    \hspace{0.1cm} \jt{Sparse symbols: $\mathcal{S}_c, \mathcal{S}_s$}; \jt{Cached output: $\widetilde O$}; \\ 
    
    \KwOut{${O}$}
    
    \SetKwFor{For}{for}{}{}  
    \SetKwFor{While}{while}{}{}
    \SetKwIF{If}{ElseIf}{Else}{if}{then}{else if}{else}{}  
    
    \SetKwFunction{sparseattn}{$FalshOmni$}
    \SetKwProg{Fn}{Func}{\string :}{}

        \For{$i \gets 1$ \KwTo $T_Q$}{
        \If{\jt{$\mathcal{F}(\mathcal{S}_c, i)==0$}}{
            {\LineComment{Cache-then-Reuse.}}
            
            {\jt{Continue if $O_i$ is already updated;}} \\
             {\jt{Load $ \widetilde O_i$ into the SM ;}} \\
             {\jt{$O_i$ = $\mathrm{OP_{reuse}}$  $(\widetilde O_i)$;}}
        } 
        \Else {
            
            {\jt{Load $ O^{t}_i$ into the SM;}}
            
            \For{$j \gets 1$ \KwTo $T_{kv}$} {
                \If{\jt{$\mathcal{J}(\mathcal{S}_s, i,j)==1$}}{
                    $\LineComment{Compute-on-Demand.}$
                    
                    {Load $ K_j$, $V_j$;}
                    
                    {$Q_iK_j^T$;} 
                    
                    {Update $m_{i,j}$, $\widetilde P_{ij}$, $l_{i,j}$;} 

                    {$\widetilde P_{ij}V_j$ $\rightarrow$ $O_{i,j}$;}
                }
            }
            
            {$O_i = \mathrm{diag}(l_{i,T_n})^{-1} O_{i,T_n}$}
        }
        {Write $O_i$ into HBM;}
    }
    \Return{$O = \{O_i\}$}\;

\end{algorithm}

\end{minipage}
\vspace{-10pt}
\end{wrapfigure}
\subsection{\method Attention}  
\looseness=-1  
Following the generation of sparse symbols \(\mathcal{S}_{c}\) and \(\mathcal{S}_{s}\), the objective of \method is to implement a general attention kernel that can interpret these symbols and execute arbitrary sparse computations efficiently. In the conventional FlashAttention framework, each CTA processes a complete computation tile. To support sparsity, we extend this design so that each CTA can detect and adapt to different sparsity patterns, as outlined in Algorithm~\ref{flashomni_attention}.  Specifically, before loading \(Q_i\), each CTA invokes the spatial‑axis decoding function \(\mathcal{F}\) (Line 5) and checks \(\mathcal{S}_{c}\) to determine whether the current tile requires computation. Depending on this result, execution follows either a compute‑on‑demand or a cache‑then‑reuse path. The decoding operation uses only bitwise procedures, expressed as  \mbox{\(\mathcal{F}(\mathcal{S}_{c}, i) = (\mathcal{S}_{c} >> i/n) ~\&~ 1\)}.

\looseness=-1  
In the cache‑then‑reuse path, element‑wise computation logic is fused into the \method attention process, allowing each CTA to choose between standard attention or lightweight element‑wise operations (e.g., summation and multiplication in TaylorSeer) based on the decoding result. Alternatively, an elementwise kernel can be invoked to perform simple value-reuse computations and write the results into the output tensor \(O\). The corresponding CTA of feature caching can then directly return (Line 7) without further processing. In the compute‑on‑demand path, an additional reduction‑axis decoding function \(\mathcal{J}\) (Line 13) is applied before updating \(O_{i,j}\) in the inner loop to determine whether to skip specific blocks by decoding \(\mathcal{S}_{s}\). This operation is defined as  \mbox{\(\mathcal{J}(\mathcal{S}_{s}, i, j) = (\mathcal{S}_{s} >> \frac{i}{n}\frac{T_{kv}}{n} + \frac{j}{n}) ~ \& ~ 1\)}. Directly performing this bitwise operation for every \(K_{j}\) load increases CUDA core overhead. To mitigate this, undecoded bits are processed only once when first encountered, and the results—covering up to \(8n\) consecutive blocks—are stored in registers for subsequent reuse.

\subsection{\method Sparse GEMMs}  
\looseness=-1  
FlashOmni introduces two \textit{sparse GEMMs}, GEMM-\(Q\) and GEMM-\(O\),  utilizing \(\mathcal{S}_{c}\) to remove redundant computations in the Linear layers: query projection (\(\mathrm{Proj_{to\_q}}\)) and output projection (\(\mathrm{Proj_{to\_out}}\)) within attention modules.  

\looseness=-1  
\texttt{Observation 2.} In typical attention modules of common DiT architectures, the computation from the input tensor \(X\) to the final attention output \(O\) involves three main stages: (1) generating the query vectors \(Q\) from \(X\); (2) using \(Q\) to compute attention scores with the corresponding keys and aggregating the values; and (3) writing the aggregated results into the output tensor \(O\). Specifically, the generation of \(Q\) from \(X\) typically consists of three sequential operations: query projection (\(\mathrm{Proj_{to\_q}}\)), token‑wise RMS normalization (\(\mathrm{OP_{RMSNorm}}\)), and rotary positional encoding (\(\mathrm{OP_{RoPE}}\)). The standard attention output formulation \mbox{\(O^{h}_{i} = \sum_{j} P^{h}_{i,j} V^{h}_{j}\)} can be expanded as Equation~\ref{gemm_q_ob}.

\vspace{-10pt}
\begin{equation}
\label{gemm_q_ob}
O^{h}_i = \sum_{j} \mathrm{Softmax} \left( \frac{\mathrm{OP_{RoPE}}\left( \mathrm{OP_{RMSNorm}}\left(X_{i} W^{h}_{to\_q}\right) \right) (K^{h})^{T}}{\sqrt{d}} \right)_{j} V^{h}_{j}
\end{equation}
\vspace{-10pt}

\looseness=-1  
Both RMSNorm and RoPE operate exclusively along the feature dimension for each token, without cross‑token computation. Therefore, at the \textit{Dispatch} step, if \(\mathcal{S}_{c}\) specifies that a particular block \(O^{h}_{i}\) is retrieved from the cache \(\widetilde{O}^{h}_{i}\), the corresponding query projection \(Q^h_i = X_i W^{h}_{to\_q}\) can be skipped.

\looseness=-1  
\textit{FlashOmni GEMM-\(Q\)}. As shown in the left part of Figure~\ref{fig:FlashOmni_overview} (\method GEMM-\(Q\)), newly computed \(Q\) values are required to refresh the symbol information at the \textit{Update} step, and the GEMM-\(Q\) operation will follow its full-standard execution. At the \textit{Dispatch} step, each CTA applies the spatial‑axis decoding function \(\mathcal{F}\) to \(\mathcal{S}_{c}\) to determine whether its block tile participates in the upcoming attention computation. If not, the CTA exits immediately without performing any further operations.

\looseness=-1  
\texttt{Observation 3.} In typical DiT architectures, once the attention output for each head, \(O^{h}_{i}\), is obtained, an output projection (\(\mathrm{Proj_{to\_out}}\)) is usually applied to facilitate information exchange across heads for the same token, which can be expressed as Equation~\ref{outlinear}, where the set \(H_i = \{ h \mid J(\mathcal{S}_{c}, i, h) = 1 \}\) denotes the heads whose attention outputs are generated directly by attention computation at the current step, rather than obtained via feature caching.    

\vspace{-10pt}
\begin{equation}
\label{outlinear}
Out_i = \sum_h O^{h}_{i} W_{to\_out}^h 
= ( \sum_{h \notin H_i} + \sum_{h \in H_i} ) O^{h}_{i} W_{to\_out}^h
\end{equation}
\vspace{-10pt}

\looseness=-1  
At the \textit{Dispatch} step, when \(\mathcal{S}_{c}\) specifies that the output features of head \(h\) for block \(i\) are retrieved from cache \(\widetilde{O}^{h}_{i}\), the operation can be expressed as \(O^{h}_{i} = \mathrm{OP_{reuse}}(\widetilde{O}^{h}_{i})\) for \(h \notin H_i\). Since \(\mathrm{OP_{reuse}}\) is an element‑wise operation, the following property holds:

\vspace{-10pt}
\begin{align}
\label{outlinear_change}
\sum_{h \notin H_i} O^{h}_{i} W_{to\_out}^h
&= \sum_{h \notin H_i} \mathrm{OP_{reuse}}(\widetilde{O}^{h}_{i}) W_{to\_out}^h = \mathrm{OP_{reuse}}( \sum_{h \notin H_i} \widetilde{O}^{h}_{i} W_{to\_out}^h ).
\end{align}
\vspace{-10pt}

\looseness=-1  
This property enables caching of \(\sum_{h \notin H_i} \widetilde{O}^{h}_{i} W_{to\_out}^h\) as a bias term \(\mathcal{B}_{c}\) at the \textit{Update} step. At later \textit{Dispatch} steps, the cached bias can be transformed using an element‑wise kernel, \(\mathrm{OP_{reuse}}(\mathcal{B}_{c})\), and added directly to remaining computations of \(\mathrm{Proj_{to\_out}}\) (\mbox{\(\sum_{h \in H_i}O^{h}_{i} W_{to\_out}^h\)}) as a bias. This design eliminates redundant reduction‑axis computations in the GEMM kernel. Furthermore, storing the cache bias removes the need to retain \(\widetilde{O}^{h}_{i}\) in HBM at \textit{Update} steps. The related element‑wise operations in \method Attention can also be skipped entirely, allowing the cache‑then‑reuse branch to terminate immediately. This design reduces both computational cost and memory consumption.

\looseness=-1
\textit{FlashOmni GEMM-\(O\)}. As illustrated in the right part of Figure~\ref{fig:FlashOmni_overview} (\method GEMM-\(O\)), the GEMM-\(O\) computation is divided into two stages. For each block tile \(O^{h}_{i}\) along the reduction axis, each CTA applies the reduction-axis decoding function \(\mathcal{J}\) on \(\mathcal{S}_{c}\) to determine whether the tile should be generated via cache‑and‑reuse or computed on demand.  At the \textit{Update} step, the latest \(\mathcal{S}_{c}\) and attention outputs \(O\) are generated, and the cache bias \(\mathcal{B}_{c}\) is subsequently refreshed. In the first stage, CTAs identify tiles that will be reused at the \textit{Dispatch} step, compute their outputs, and record the results in \(\mathcal{B}_{c}\). The second stage then relaunches the kernel, processing the remaining tiles that are always updated through attention computations at the next future steps and adding the cached bias. At the \textit{Dispatch} step, the GEMM-\(O\) output space is initialized via \(\mathrm{OP_{reuse}}\), executing only the computations corresponding to the second stage of the \textit{Update} step. This strategy eliminates redundant operations on cached tiles, thereby reducing both computational workload and memory usage.

\section{Experiments}

\subsection{Setup}
\looseness=-1
\textbf{Models}. Our experiments are performed on three SOTA visual generative models: the text‑to‑image generator FLUX.1‑dev~\citep{flux1}, the text‑to‑video generator HunyuanVideo~\citep{kong2024hunyuanvideo}, and the text‑guided image editing model FLUX.1‑Kontext~\citep{batifol2025flux}.

\looseness=-1
\textbf{Benchmarks and Metrics}.  
For the text‑to‑image generation experiments, we conduct inference on the COCO‑2017 dataset~\citep{lin2014microsoft} at a resolution of \(1024 \times 1024\) pixels. The text‑guided image editing evaluation is performed on KontextBench~\citep{batifol2025flux}, which contains 1,026 unique image–prompt pairs derived from 108 base images. Generated outputs are assessed using CLIP‑IQA~\citep{wang2023exploring} and FID‑FP16~\citep{heusel2017gans}, measuring image quality and distributional similarity, respectively. For text‑to‑video generation, we adopt VBench~\citep{huang2024vbench} to evaluate various aspects of video quality, including background consistency, motion smoothness, temporal flicker, and stylistic coherence.   Fidelity to the original results is quantified using PSNR, SSIM~\citep{wang2004image}, and LPIPS~\citep{zhang2018unreasonable}.

\looseness=-1
To measure computational efficiency of sparse attention, we report TOPS (tera‑operations per second), defined as \(\texttt{attn}/\texttt{t}\), and \textbf{Sparsity}, defined as (\texttt{skip} / \texttt{total}) following SpargeAttn~\citep{zhangspargeattention}. Here, \(\texttt{attn}\) denotes the total number of operations in a standard attention computation, \(\texttt{t}\) is the latency from given \((Q, K, V)\) inputs to the corresponding attention outputs, \texttt{skip} is the number of skipped attention pairs \((Q_iK_j^\top, \widetilde{P}_{ij}V_j)\), and \texttt{total} is the total number of such pairs.

\looseness=-1
\textbf{Baselines and Platform}. We compare \method with five SOTA baselines: two block‑sparse attention skipping methods: SpargeAttn~\citep{zhangspargeattention} and DiTFastAttnV2~\citep{zhang2025ditfastattnv2}, and three feature caching methods: ToCa~\citep{zouaccelerating}, FORA~\citep{selvaraju2024fora}, and TaylorSeer~\citep{liu2025reusing}. The configuration for \method is specified as \((\tau_{q}, \tau_{kv}, \mathcal{N}, \mathcal{D}, S_q)\) (Appendix~\ref{setup_for_hyper}). All experiments are executed on a single NVIDIA A100 GPU.

\renewcommand \arraystretch{1.}
\begin{table}[t]
    \caption{
    	 End-to-end metrics comparison with block-sparse skipping across image and video generation models.
    }
    \vspace{-5pt}
    \label{exp:e2e_bss}

    \scriptsize \centering
    \resizebox{\textwidth}{!}{
    \begin{tabular}{l|l|ccccccccc}
    \toprule
    {\textbf{Method}(seq\_len)} & \multirow{2}{*}{\bf Configuration}  & \multirow{2}{*}{\bf TOPS ($\uparrow$)} & \multirow{2}{*}{\bf Sparsity (\%, $\uparrow$)}  & \multirow{2}{*}{\bf PSNR ($\uparrow$)}& \multirow{2}{*}{\bf LPIPS ($\downarrow$) } & \multirow{2}{*}{\bf SSIM $\uparrow$} & \multirow{2}{*}{\bf CLIP-IQA ($\uparrow$)} & \multirow{2}{*}{\bf FID ($\downarrow$)}\\
    {\bf FLUX.1(4.5K)} &  &  & &  &  & &  \\
    
    \midrule
     \multirow{2}{*}{Full-Attention}  &  {[dev]: 50 steps}  & \multirow{2}{*}{93.82} & 0 & $\infty$ & ---  & --- & 0.5154 & --- \\
    &    {50\% steps}  & & 50 & 16.69 & 0.3426 & 0.7004 & 0.5054  &  56.589 \\
    \cmidrule{1-9}

    {DiTFastAttnV2} & ($\theta=0.2$) & 114.25 & 26 & 20.289 & 0.2174 & 0.7901 & 0.5037 &  39.436 \\
    {SpargeAttn} &($l_1=6.5\%, l_2=7\%$) & 101.41 &  22 & 21.358 & 0.1868 & 0.8148 & 0.5072 & 34.156 \\
    \cellcolor{gray!20}{Dyn-Sparse} & \cellcolor{gray!20}($5\%, 15\%, 4, 0, 0\%$)& \cellcolor{gray!20}133.12  & \cellcolor{gray!20}31  & \cellcolor{gray!20}24.259 &\cellcolor{gray!20}0.1209 & \cellcolor{gray!20}0.8665  & \cellcolor{gray!20}0.5129  &  \cellcolor{gray!20}25.163  \\
    
    \cellcolor{gray!20}{\method} & \cellcolor{gray!20}($5\%, 15\%, 4, 0, 0\%$) &\cellcolor{gray!20}119.35   &\cellcolor{gray!20}28 &\cellcolor{gray!20}24.563 &\cellcolor{gray!20}0.1154   &\cellcolor{gray!20}0.8716   &\cellcolor{gray!20}0.5124   &\cellcolor{gray!20}23.942  \\
    
    {\cellcolor{gray!20}\method} & \cellcolor{gray!20}($50\%, 15\%, 4, 1, 0\%$) &\cellcolor{gray!20}149.25 & \cellcolor{gray!20}41 & \cellcolor{gray!20}\textbf{25.159} & \cellcolor{gray!20}\textbf{0.0992} & \cellcolor{gray!20}\textbf{0.8838} & \cellcolor{gray!20}0.5126 &   \cellcolor{gray!20}\textbf{20.933} \\
    
    {\cellcolor{gray!20}\method} & \cellcolor{gray!20}($50\%, 15\%, 5, 1, 0\%$) &\cellcolor{gray!20}152.25 & \cellcolor{gray!20}43 & \cellcolor{gray!20}23.994 & \cellcolor{gray!20}0.1241 & \cellcolor{gray!20}0.8618 & \cellcolor{gray!20}\textbf{0.5165} &   \cellcolor{gray!20}25.594 \\
    
    {\cellcolor{gray!20}\method} & \cellcolor{gray!20}($50\%, 15\%, 5, 2, 30\%$) &\cellcolor{gray!20}\textbf{163.19} & \cellcolor{gray!20}\textbf{46} &\cellcolor{gray!20}23.859 & \cellcolor{gray!20}0.1281 & \cellcolor{gray!20}0.8575 & \cellcolor{gray!20}0.5123  & \cellcolor{gray!20}25.997 \\
    
    \bottomrule
    \end{tabular}
    }


    \vspace{2pt}
    \centering
    \resizebox{\textwidth}{!}{
    \begin{tabular}{l|cccccccccc}
    \toprule
    {\textbf{Method}(seq\_len)} & \multirow{2}{*}{\bf TOPS ($\uparrow$)} & \multirow{2}{*}{\bf Sparsity($\uparrow$)}  & \multirow{2}{*}{\bf PSNR ($\uparrow$)}& \multirow{2}{*}{\bf LPIPS ($\downarrow$) } & \multirow{2}{*}{\bf SSIM ($\uparrow$)} & \multirow{2}{*}{\bf Smoothness ($\uparrow$)} & \multirow{2}{*}{\bf Consistency ($\uparrow$) } & \multirow{2}{*}{\bf Flickering ($\uparrow$) } & \multirow{2}{*}{\bf Style ($\uparrow$) } \\
    {\bf Hunyuan Video (33K)} &   & &  &  & & &\\
    
    \midrule
     {Full-Attention (50 steps)}   &  92.32 & 0 & $\infty$ & --- & --- & 99.33 & 97.87 & 99.221 & 0.2394 \\
    \cmidrule{1-10}

    {DiTFastAttnV2 ($\theta=0.2$)}& 124.83 & 31 & 19.751 & 0.3086 & 0.7331 & 99.13& 97.44 & 99.098 & 0.2381 \\
    {SpargeAttn ($l_1=6\%, l_2=6.5\%$)} & 110.54 & 32 & 21.701 & 0.2442 & 0.7748 & 99.21 & 97.42 & 98.993 & 0.2399 \\
    
    {\cellcolor{gray!20}\method ($40\%, 1\%, 3, 1, 0$)}   & \cellcolor{gray!20}135.93 &\cellcolor{gray!20}34 & \cellcolor{gray!20}\textbf{32.192} & \cellcolor{gray!20}\textbf{0.0571} & \cellcolor{gray!20}\textbf{0.9289} & \cellcolor{gray!20}\textbf{99.32} &  \cellcolor{gray!20}\textbf{97.69} &  \cellcolor{gray!20}\textbf{99.217}  &  \cellcolor{gray!20}\textbf{0.2401}\\
    
    {\cellcolor{gray!20}\method ($40\%, 1\%, 6, 1, 0$)} & \cellcolor{gray!20}149.85 &\cellcolor{gray!20}39 & \cellcolor{gray!20}28.733 & \cellcolor{gray!20}0.0946 & \cellcolor{gray!20}0.8918 & \cellcolor{gray!20}99.29 &  \cellcolor{gray!20}97.55 &  \cellcolor{gray!20}99.198  &  \cellcolor{gray!20}\textbf{0.2401} \\
    
    {\cellcolor{gray!20}\method ($50\%, 5\%, 6, 1, 30\%$)} & \cellcolor{gray!20}\textbf{173.12} &\cellcolor{gray!20}\textbf{47}  & \cellcolor{gray!20}27.877 & \cellcolor{gray!20}0.1195 & \cellcolor{gray!20}0.8699 & \cellcolor{gray!20}\textbf{99.33} &  \cellcolor{gray!20}97.55 &  \cellcolor{gray!20}\textbf{99.224}&  \cellcolor{gray!20}0.2392\\
    
    \bottomrule
    \end{tabular}
    }
    
\end{table}

\renewcommand \arraystretch{1.}
\begin{table}[t]
    \caption{
    	 End-to-end metrics comparison with feature caching across image and video generation models.
    }
    \vspace{-5pt}
    \label{exp:e2e_fc}

    \scriptsize \centering
    \resizebox{\textwidth}{!}{
    \begin{tabular}{l|l|ccccccc}
    \toprule
    {\textbf{Method}(seq\_len)} & \multirow{2}{*}{\bf Configuration}  & \multirow{2}{*}{\bf PSNR ($\uparrow$)}& \multirow{2}{*}{\bf LPIPS ($\downarrow$) } & \multirow{2}{*}{\bf SSIM $\uparrow$} & \multirow{2}{*}{\bf CLIP-IQA ($\uparrow$)} & \multirow{2}{*}{\bf FID ($\downarrow$)}\\
    {\bf FLUX.1(4.5K)} &  &  &  & &  \\
    
    \midrule
     \multirow{2}{*}{Full-Attention}  &  {[dev]: 50 steps}  & $\infty$ & ---  & --- & 0.5154 & --- \\
    &    {50\% steps}   & 16.69 & 0.3426 & 0.7004 & 0.5054  &  56.589 \\
    \cmidrule{1-7}
    
    {FORA} & \multirow{2}{*}{($\mathcal{N}=5$)}   & 22.846 & 0.1595 & 0.825 & 0.5111 & 31.188 \\
    {ToCa} &   & 22.827 & 0.2059 &  0.7978 & 0.5095  & 29.947 \\
    {TaylorSeer} & ($\mathcal{N}=5,\mathcal{D}=1$)   & 22.852 & 0.1441 & 0.8361 & 0.5103 &  28.253 \\
    {TaylorSeer} & ($\mathcal{N}=5,\mathcal{D}=2$)  & 23.177 & 0.1402 & 0.8395 & 0.5131 & 27.376 \\

    {\cellcolor{gray!20}\method} & \cellcolor{gray!20}($50\%, 15\%, 5, 0, 30\%$)  & \cellcolor{gray!20}23.355 & \cellcolor{gray!20}0.1452 & \cellcolor{gray!20}0.8452 & \cellcolor{gray!20}0.5159 &  \cellcolor{gray!20}28.925 \\
    
    {\cellcolor{gray!20}\method} & \cellcolor{gray!20}($50\%, 15\%, 5, 1, 30\%$) &\cellcolor{gray!20}23.866 & \cellcolor{gray!20}0.1283 & \cellcolor{gray!20}0.8575 & \cellcolor{gray!20}0.5119  & \cellcolor{gray!20}25.994 \\

    {\cellcolor{gray!20}\method} & \cellcolor{gray!20}($50\%, 15\%, 5, 2, 30\%$) &\cellcolor{gray!20}23.859 & \cellcolor{gray!20}0.1281 & \cellcolor{gray!20}0.8575 & \cellcolor{gray!20}0.5123  & \cellcolor{gray!20}25.997 \\
    
    {\cellcolor{gray!20}\method} & \cellcolor{gray!20}($50\%, 15\%, 5, 1, 0\%$) & \cellcolor{gray!20}\textbf{23.994} & \cellcolor{gray!20}\textbf{0.1241} & \cellcolor{gray!20}\textbf{0.8618} & \cellcolor{gray!20}\textbf{0.5165} &   \cellcolor{gray!20}\textbf{25.594} \\
    
    \cmidrule{1-7}
    {TaylorSeer} & ($\mathcal{N}=6,\mathcal{D}=2$)  & 22.245 & 0.1615 & 0.8199 & 0.5119   & 31.082 \\
    
    {\cellcolor{gray!20}\method} & \cellcolor{gray!20}($50\%, 15\%, 6, 1, 30\%$)   & \cellcolor{gray!20}\textbf{23.217}  & \cellcolor{gray!20}\textbf{0.1507} & \cellcolor{gray!20}\textbf{0.8373} & \cellcolor{gray!20}\textbf{0.5124} & \cellcolor{gray!20}\textbf{28.965} \\
    \bottomrule
    \end{tabular}
    }

    \vspace{2pt}
    \centering
    \resizebox{\textwidth}{!}{
    \begin{tabular}{l|cccccccccc}
    \toprule
    {\textbf{Method}(seq\_len)}  & \multirow{2}{*}{\bf PSNR ($\uparrow$)}& \multirow{2}{*}{\bf LPIPS ($\downarrow$) } & \multirow{2}{*}{\bf SSIM ($\uparrow$)} & \multirow{2}{*}{\bf Smoothness ($\uparrow$)} & \multirow{2}{*}{\bf Consistency ($\uparrow$) } & \multirow{2}{*}{\bf Flickering ($\uparrow$) } & \multirow{2}{*}{\bf Style ($\uparrow$) } \\
    {\bf Hunyuan Video (33K)} &  &  & & &\\
    
    \midrule
     {Full-Attention (50 steps)}    & $\infty$ & --- & --- & 99.33 & 97.87 & 99.221 & 0.2394 \\
    \cmidrule{1-8}
    
    {TaylorSeer ($\mathcal{N}=6,\mathcal{D}=1$)} & 26.483 & 0.1229 & 0.8621 & 99.32& 97.38 & 99.196 & 0.2388 \\
    {\cellcolor{gray!20}\method ($50\%, 5\%, 6, 1, 30\%$)}  & \cellcolor{gray!20}27.877 & \cellcolor{gray!20}0.1195 & \cellcolor{gray!20}0.8699 & \cellcolor{gray!20}\textbf{99.33} &  \cellcolor{gray!20}97.55 &  \cellcolor{gray!20}\textbf{99.224}&  \cellcolor{gray!20}0.2392\\
    
    {\cellcolor{gray!20}\method ($40\%, 1\%, 6, 1, 0$)} & \cellcolor{gray!20}\textbf{28.733} & \cellcolor{gray!20}\textbf{0.0946} & \cellcolor{gray!20}\textbf{0.8918} & \cellcolor{gray!20}99.29 &  \cellcolor{gray!20}\textbf{97.55} &  \cellcolor{gray!20}99.198  &  \cellcolor{gray!20}\textbf{0.2401} \\
    
    \bottomrule
    \end{tabular}
    }
    
\end{table}

\subsection{Visual Quality Evaluation}
\looseness=-1
As shown in Table~\ref{exp:e2e_bss} and Table~\ref{exp:e2e_fc}, \method consistently outperforms all baselines across quality metrics on FLUX and HunyuanVideo. Compared with block‑sparse skipping (DiTFastAttnV2, SpargeAttn), it achieves higher sparsity under various threshold settings while maintaining clearly superior quality.  Against caching methods (FORA, ToCa, TaylorSeer), \method attains better quality on FLUX under the same moderate cache interval ($\mathcal{N}$ = 5) and order $\mathcal{D}$. Even under the more aggressive interval ($\mathcal{N}$ = 6), \method maintains substantial advantages on both FLUX ($1.74\uparrow$ SSIM over TaylorSeer) and HunyuanVideo (\(2.97\uparrow\) SSIM over TaylorSeer). Similar performance gains are also observed on FLUX.1-Kontext (Table~\ref{exp:kontext}), where \method surpasses all other methods.

\subsection{Efficiency Evaluation}

\begin{figure}[t]
    \centering
    \includegraphics[width=\linewidth]{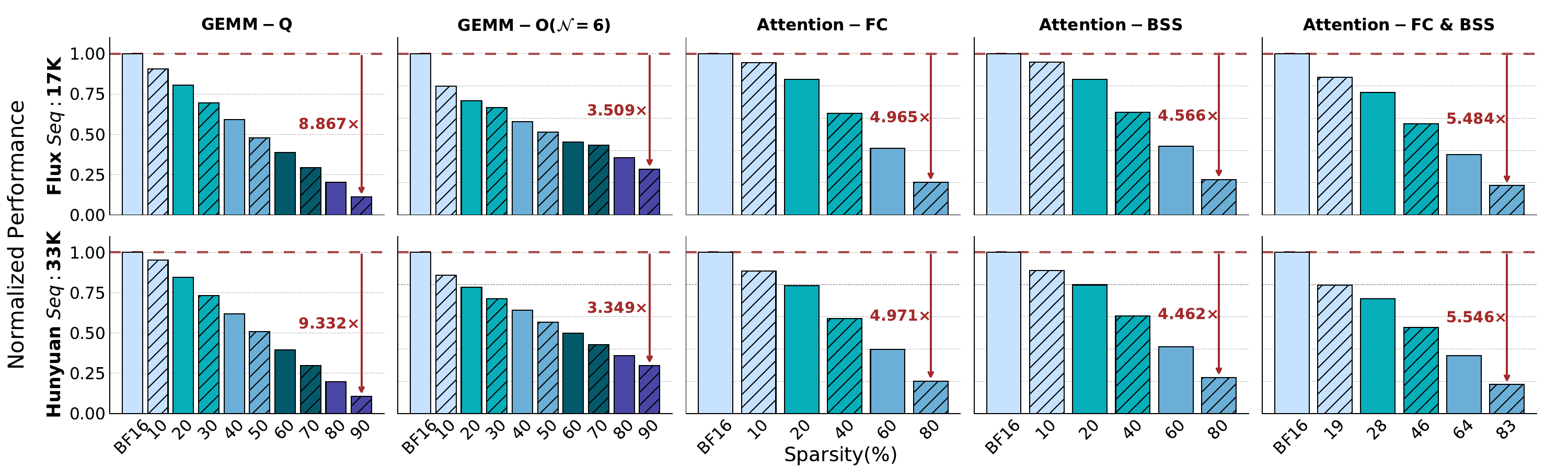}
    \vspace{-15pt}
	\caption{Normalized inference performance on NVIDIA A100 (BF16, measured by using cudaLib and FlashInfer) at different sparsity levels for sparse GEMM-$Q$/-$O$ ($\mathcal{N}=6$) and Attention kernels. FC and BSS indicate the application of feature caching and block-sparse skipping, respectively.}
    \vspace{-10pt}
    \label{exp:FlashOmni_kernel}
\end{figure}
\textbf{Attention performance.}
\looseness=-1
We implement our sparse attention kernel based on Flashinfer~\citep{ye2025flashinfer}, designed to remove computations deemed unnecessary by sparse symbols dynamically. We evaluate its performance through a systematic comparison with relevant baselines using randomly generated sparse symbols under three configurations: (1) feature caching (FC) only, (2) block-sparse skipping (BSS) only, and (3) both enabled. For each configuration, we report speedup across varying sparsity ratios and compare it with the theoretical computation reduction.  The right three columns in Figure~\ref{exp:FlashOmni_kernel} show that the empirical speedup closely matches the theoretical prediction and scales linearly with sparsity. More detailed experimental results are provided in the Appendix~\ref{appendix_attention}.

\looseness=-1
At the same sparsity level, FC consistently yields higher performance than BSS. For instance, at 80\% sparsity, FC achieves a 4.97$\times$ speedup, while BSS reaches 4.6$\times$. This difference arises because FC requires decoding only once per CTA, whereas BSS performs decoding repeatedly throughout the reduction process, incurring additional CUDA core overhead.  For video generation, as shown in Figure~\ref{exp:e2e_bss}, \method delivers up to $1.5\times$ end-to-end acceleration under a 46\% sparsity setting.

\looseness=-1
\textbf{Sparse GEMMs performance.} We also implement our sparse GEMM kernels based on CuTe to mitigate redundant computation and memory overhead introduced by enabling the FC strategy in attention. We conduct extensive experiments on both the GEMM-\(Q\)/-$O$, observing a consistent linear speedup trend with increasing sparsity. For GEMM-\(Q\),  acceleration occurs along the spatial axis. Since decoding is performed only once, the achieved speedup closely matches the theoretical upper bound of computation reduction. For GEMM-\(O\), sparsity lies along the reduction axis, requiring multiple decoding operations, preventing a one-to-one match with the theoretical peak speedup; nevertheless, the achieved performance remains close. For instance, at 90\% sparsity with $\mathcal{N}=6$, the theoretical speedup is $4\times$ (speedup analysis refers to Appendix~\ref{speedupdefination}), while our kernel attains $3.3\times\sim3.5\times$ (Figure~\ref{exp:FlashOmni_kernel}). This discrepancy arises because GEMM-\(O\) requires multiple decoding operations along the reduction axis on CUDA cores. As for a single inference, even at 90\% sparsity, while  GEMM-\(Q\) alone can achieve a near‑linear \(9\times\) speedup, GEMM-\(O\) only reaches \(6.39\times\) due to these additional overheads. More details refer to the Appendix~\ref{appendix_gemm_o}.

\begin{wrapfigure}{r}{0.4\linewidth}
    \vspace{-10pt}
    \centering
    \includegraphics[width=\linewidth]{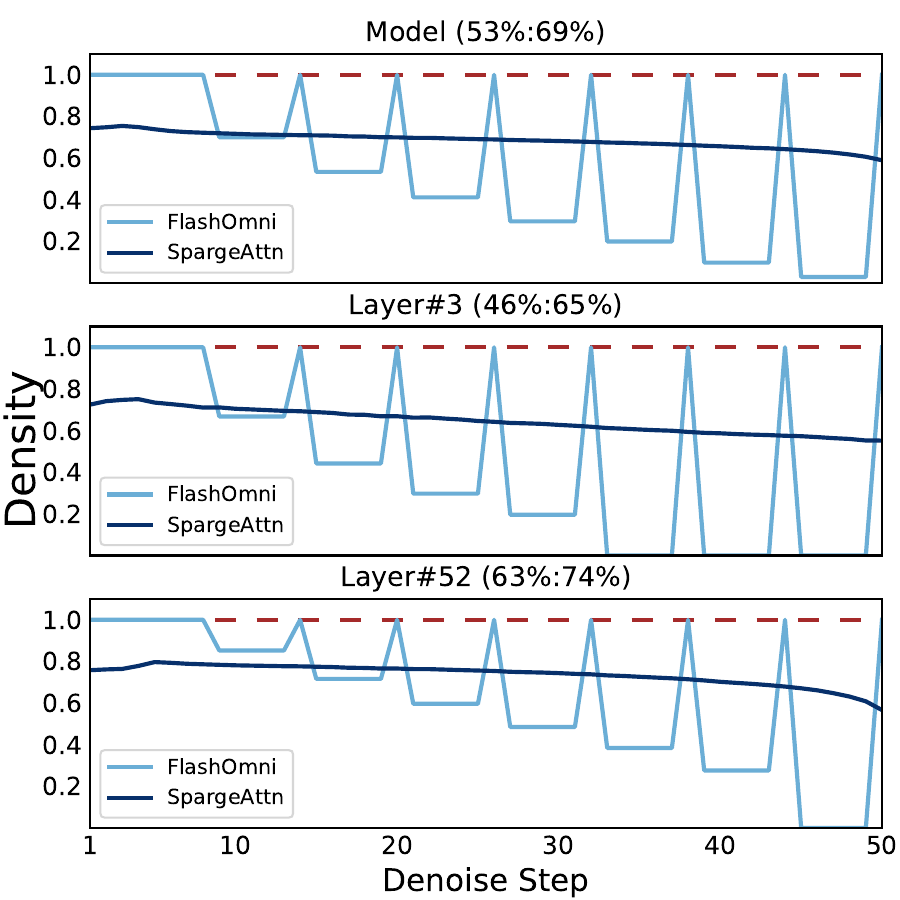}
    \vspace{-20pt}
    \caption{\looseness=-1 Density comparison with SpargeAttn on HunyuanVideo benchmark.}
    \vspace{-30pt}
    \label{fig:density}
\end{wrapfigure}

\looseness=-1
\textbf{Density analysis.} We further measure the computation density during inference on HunyuanVideo, as illustrated in Figure~\ref{fig:density}. For SpargeAttn, the density remains relatively stable throughout the inference process. In contrast, FlashOmni exhibits a sharp drop in density from an initial value close to 1, maintaining overall lower density both at the whole-model level and within individual transformer layers. We attribute \method's ability to achieve higher video generation quality despite lower density to the critical role of the early denoising steps in the DiT inference process.  In these initial stages, the vision token is randomly initialized from Gaussian noise and requires strong and comprehensive text guidance, alongside timely updates of the textual signal, consistent with our findings in Observation 1.   Moreover, under our degradation strategy, when the proportion of token blocks requiring updates falls below a certain threshold, we directly employ full-feature caching to improve efficiency, whose details refer to Appendix~\ref{setup_for_hyper}. 

\subsection{Ablation study}
\renewcommand \arraystretch{1.}
\begin{table}[t]
    \caption{
        The ablation study evaluates \method with different configurations on {FLUX.1}.
    }
    \vspace{-5pt}
    \label{exp:ablation}
    \scriptsize \centering
    \begin{tabular}{l|cccc}
    \toprule
    {\bf Configuration}  & {\bf PSNR ($\uparrow$)} & {\bf LPIPS ($\downarrow$)} & {\bf SSIM ($\uparrow$)} & {\bf FID ($\downarrow$)} \\
    \midrule
    ($5\%, 15\%, \textcolor{red}{\mathcal{N}=3}, 1, 0$)
     & 26.702 & 0.0784 & 0.9041 & 16.495 \\
    ($5\%, 15\%, \textcolor{red}{\mathcal{N}=4}, 1, 0$)
     & 25.217 & 0.0999 & 0.8837 & 20.789 \\
    ($5\%, 15\%, \textcolor{red}{\mathcal{N}=5}, 1, 0$)
     & 24.193 & 0.1214 & 0.8650 & 24.860 \\
    ($5\%, 15\%, \textcolor{red}{\mathcal{N}=6}, 1, 0$)
     & 23.454 & 0.1366 & 0.8506 & 28.072 \\
    ($5\%, 15\%, \textcolor{red}{\mathcal{N}=7}, 1, 0$)
     & 23.012 & 0.1488 & 0.8405 & 30.541 \\
    \cmidrule{1-5}
   ($50\%, 15\%, 5, \textcolor{red}{\mathcal{D}=0}, 30\%$)
    & 23.355 & 0.1452 & 0.8452 & 28.925\\
   ($50\%, 15\%, 5, \textcolor{red}{\mathcal{D}=1}, 30\%$)
     & 23.866 & 0.1283 & 0.8575 & 25.994 \\
   ($50\%, 15\%, 5, \textcolor{red}{\mathcal{D}=2}, 30\%$)
     & 23.859 & 0.1281 & 0.8575 & 25.997 \\
    \bottomrule
    \end{tabular}

\end{table}

\begin{figure}[t]
    \centering
    \includegraphics[width=\linewidth]{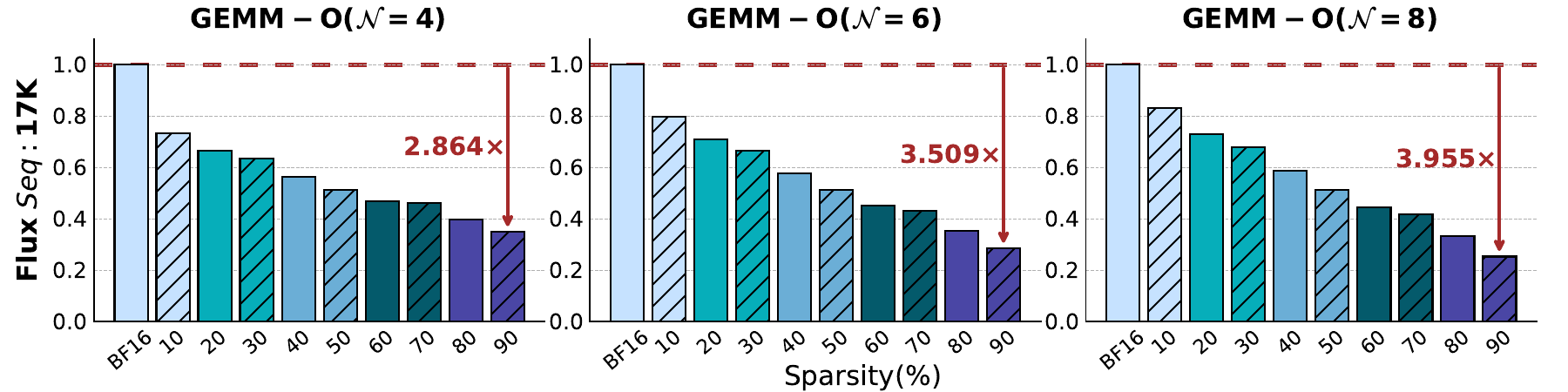}
    \vspace{-15pt}
	\caption{Normalized inference performance at different \(\mathcal{N}\) for sparse GEMM-$O$, where token length is 17K.}
    \vspace{-10pt}
    \label{exp:FlashOmni_out_33k_kernel}
\end{figure}
\looseness=-1
We perform ablation experiments (Table~\ref{exp:ablation}) on Flux to assess how the interval \(\mathcal{N}\) and caching order \(\mathcal{D}\) affect generation quality. First‑order caching (\(\mathcal{O}=1\)) delivers the largest performance gains, while higher orders degrade quality, highlighting the limits of simulation in capturing real‑world behavior. Despite its bias, first‑order caching remains more faithful than direct reuse. Increasing \(\mathcal{N}\) offers greater acceleration but at the cost of noticeable quality loss.

We further conduct ablation experiments (Figure~\ref{exp:FlashOmni_out_33k_kernel}) to evaluate the speedup of GEMM-\(O\) across settings \(\mathcal{N}\) with a sequence length of 17K. As \(\mathcal{N}\) increases, the overall speedup consistently improves but falls short of the theoretical speedup predicted by reduced FLOPs. For example, when \(\mathcal{N} = 4, 6, 8\), the measured  93.1\%, 87.7\%, and 84.7\% of theoretical speedup , respectively. This discrepancy arises because GEMM-\(O\) requires multiple decoding operations along the reduction axis on CUDA cores; even at 90\% sparsity, while  GEMM-\(Q\) alone can achieve a near‑linear \(9\times\) speedup, GEMM-\(O\) only reaches \(6.39\times\) due to these additional overheads. As \(\mathcal{N}\) grows, this effect becomes increasingly pronounced in the aggregated GEMM-\(O\) speedup.  

\section{Conclusion}
\looseness=-1
We propose FlashOmni, a unified sparse attention engine that is applicable to any Diffusion Transformers. FlashOmni abstracts an \textit{Update–Dispatch} paradigm with three key designs: it unifies multi-granularity spare strategies with flexible sparse symbols, supports efficient arbitrary sparse attention through a general kernel, and optimizes sparse GEMMs in attention linear layers to remove redundant computation. Experiments demonstrate that FlashOmni achieves significant end-to-end acceleration across diverse models without retraining, while preserving generation quality.

\bibliography{main}
\bibliographystyle{flashomni_2026_conference}

\appendix
\section{Appendix}

\subsection{Experimental Details}
\subsubsection{Setup}
\label{setup_for_hyper}

\looseness=-1
As is metioned in 4.1, the configuration for \method is specified as \((\tau_{q}, \tau_{kv}, \mathcal{N}, \mathcal{D}, S_q)\). In Table~\ref{exp:fulubiao1}, we give details about them.

\begin{table}[h!]
\scriptsize \centering
\renewcommand{\arraystretch}{1.3}
\begin{tabular}{|>{\centering\arraybackslash}m{2cm}|p{8cm}|} 
\hline
\textbf{Settings} & \multicolumn{1}{c|}{\textbf{Description}} \\
\hline
\raisebox{-1.5\totalheight}{$\tau_q$} & Sparsity threshold for $q$. 
Details: the importance scores of tokens are sorted in ascending order, and 
tokens are progressively marked for sparsification until the cumulative importance 
of the selected tokens exceeds $\tau_q$. \\
\hline
\raisebox{-1.5\totalheight}{$\tau_{kv}$} & Sparsity threshold for kv. Details: the importance scores of blocks are sorted in ascending order, and blocks are progressively marked for sparsification until the cumulative importance of the selected blocks exceeds $\tau_{kv}$. \\
\hline
$\mathcal{N}$& Moderate cache interval.\\
\hline
$\mathcal{D}$& Order of expansion.\\
\hline
\raisebox{-0.7\totalheight}{$S_q$}&Threshold of caching. Details: if the proportion of tokens requiring computation is below this threshold, the layer degenerates into feature caching.\\
\hline
\end{tabular}
\caption{Settings and their descriptions.}\label{exp:fulubiao1}
\end{table}

It is worth noting that the values of $\tau_{q}$ and $\tau_{kv}$ are not set to their target values at the initial time step. Rather, they progressively converge to these values as the time step advances.

\looseness=-1
We conducted experiments on three models for different tasks: FLUX for the text-to-image generator, HunyuanVideo for the text-to-video generator, and FLUX.1-Kontext for text-guided image editing. We selected five representative parameters for evaluation, including $\tau_{q}$ at 5\% and 50\%; $\tau_{kv}$ at 15\%, $\mathcal{N}$ of 3, 4, 5, 6, and 7; $\mathcal{D}$ of 0, 1, and 2; $S_q$ at 0\% and 30\%. Our experiments achieved excellent results, maintaining high generation quality even under high sparsity levels. It is worth noting that these parameters can be efficiently tuned via lightweight search algorithms to further enhance the performance of \method. We plan to implement this optimization in future work.

\subsubsection{GEMM-\(O\) Speedup Metrics}
\label{speedupdefination}

\looseness=-1
In this section, we define our methodology for computing the speedup metric of GEMM-\(O\). Define $T_{total}$ as the total time taken to execute a single \(\mathrm{Proj_{to\_out}}\) operation  and $s$ to denote the sparsity ratio, then we can compute the normal time consumption: $\mathcal{N}T_{total}$, and the \method time consumption: 
\vspace{-3pt}
\begin{align}
T_{\text{FlashOmini}} &= T_{\text{sparse}} + T_{\text{computation}} \notag \\
&= T_{\text{total}}\times s + \sum_{\mathcal{N}} \left( (1-s) \times T_{\text{total}}\right) \notag \\
&= T_{\text{total}}+(\mathcal{N}-1)(1-s)T_{\text{total}}
\end{align}
Then we can compute the speedup metric $\frac{\mathcal{N}}{1+(\mathcal{N}-1)(1-s)} $. When the $s=0.9$ and $\mathcal{N}=6$, the theoretical speedup is $\frac{6}{1+(6-1)(1-0.9)}=4 $. As is shown in Figure~\ref {exp:FlashOmni_kernel}, our actual speedup obtained under this configuration is \(3.509\times\), approaching the theoretical value of \(4\times\), which substantiates the high efficiency of \method.

\subsubsection{Warmup Steps Analysis for FLUX}
\begin{figure}
    \centering
    \includegraphics[width=\linewidth]{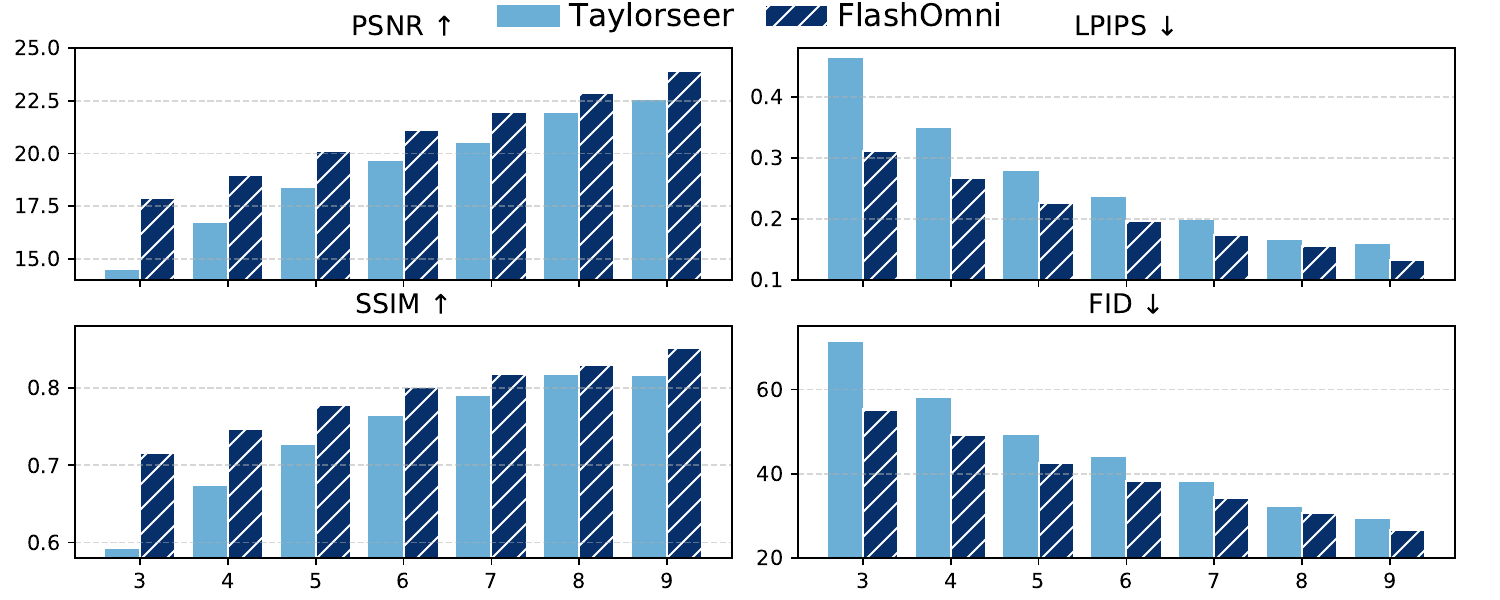}
    \label{exp:kernel_performance}
    \vspace{-20pt}
    \caption{\looseness=-1 End-to-end metrics comparison with feature caching on FLUX with different warmup steps.}
    \vspace{-5pt}
    \label{exp:vs_taylorseer}
\end{figure}

\begin{table}[h]
\scriptsize
    \centering
    \captionof{table}{
        End-to-end metrics comparison on text-guided image editing model.
    }
    \label{exp:kontext}
    \resizebox{0.8\textwidth}{!}{
    \begin{tabular}{l|l|cccc}
    \toprule
    {\textbf{Method}} & \multirow{2}{*}{\bf Configuration}   & \multirow{2}{*}{\bf PSNR ($\uparrow$)}& \multirow{2}{*}{\bf LPIPS ($\downarrow$) } & \multirow{2}{*}{\bf SSIM $\uparrow$}  & \multirow{2}{*}{\bf FID ($\downarrow$)}\\
    {\bf FLUX.1-Kontext}  & &  &  & \\
    
    \midrule
     {Full-Attention}  &  {50 steps}  & $\infty$ & --- & --- & --  \\
    \cmidrule{1-6}

    {DiTFastAttnV2} & ($\theta=0.2$) &  24.507 & 0.1233 & 0.8225 & 37.232 \\
    {SpargeAttn} &($l_1=6\%, l_2=6.5\%$)  &  26.851 & 0.1048 &  0.8519  & 28.163  \\

    {\cellcolor{gray!20}\method} & \cellcolor{gray!20}($50\%, 15\%, 5, 1, 0$)  & \cellcolor{gray!20}\textbf{31.590}  & \cellcolor{gray!20}\textbf{0.0466} & \cellcolor{gray!20}\textbf{0.9201} &  \cellcolor{gray!20}\textbf{13.218} \\
    
    \cmidrule{1-6}
    {TaylorSeer} & ($\mathcal{N}=5,\mathcal{D}=1$) & 29.733 & 0.062 & 0.8919  &  15.876 \\
    
    {\cellcolor{gray!20}\method} & \cellcolor{gray!20}($50\%, 15\%, 5, 1, 20\%$)  & \cellcolor{gray!20}\textbf{30.689}  & \cellcolor{gray!20}\textbf{0.0543} & \cellcolor{gray!20}\textbf{0.9082} & \cellcolor{gray!20}\textbf{15.204}\\
    \bottomrule
    \end{tabular}
    }
\end{table}

\looseness=-1
Figure~\ref {exp:vs_taylorseer} shows the end-to-end metrics comparison with feature caching on FLUX with different warmup steps. It is evident that when the warmup step count is low, the TaylorSeer method experiences a significant drop in image quality, with poor performance across PSNR, LPIPS, SSIM, and FID metrics, indicating a strong dependence on a high number of warmup steps. In contrast, while the \method method also shows a decline in quality at lower warmup steps, it still maintains relatively high generation quality and does not require a large warmup step count to achieve satisfactory results.

\subsubsection{Supplementary FLUX.1-Kontext Results}
\label{appendix_visualization}
Table~\ref{exp:kontext} complements the evaluation by providing an end-to-end metrics comparison on a text-guided image editing model. The results demonstrate the superior performance of \method in the FLUX.1-Kontext setting for text-guided image editing.

\subsection{\method Attention}
\label{appendix_attention}
\begin{figure}[t]
    \centering
    \includegraphics[width=\linewidth]{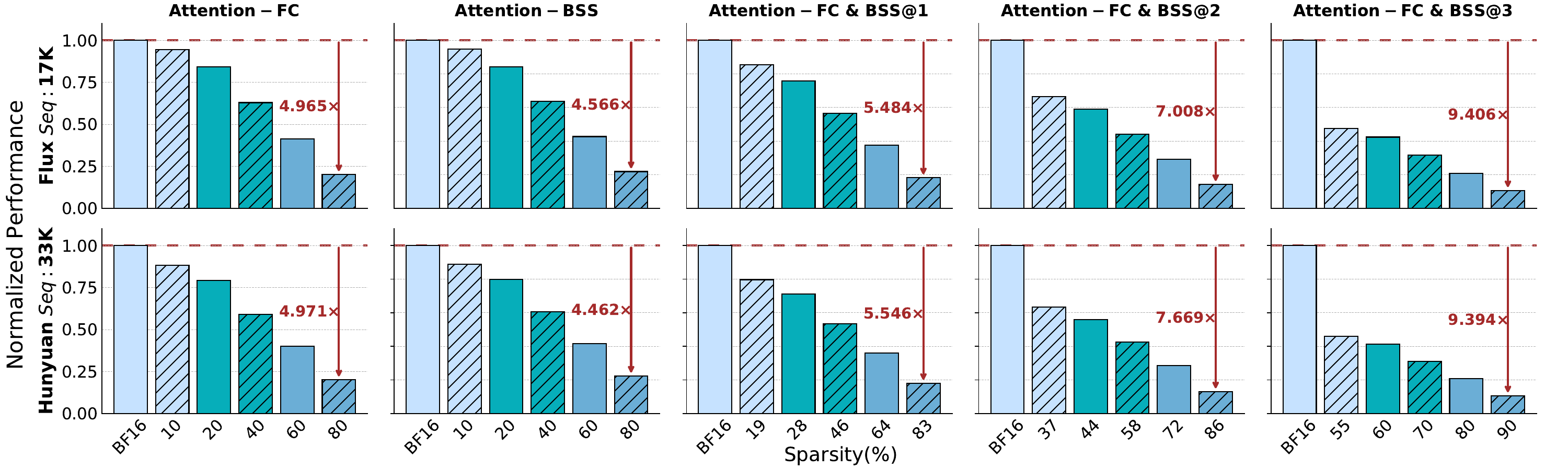}
    \vspace{-15pt}
	\caption{Normalized inference performance on NVIDIA A100 (BF16, measured by using FlashInfer) at different sparsity levels for sparse attention kernels. FC and BSS indicate the application of feature caching and block-sparse skipping, respectively.}
    \vspace{-10pt}
    \label{exp:FlashOmni_attn_kernel}
\end{figure}

\looseness=-1
We evaluate the inference performance of the attention kernel under different sparsity levels during high-resolution diffusion, including 2K image generation in Flux, video generation in HunyuanVideo, using token lengths of 17K and 33K as examples. Three configurations are tested: activating only FC, activating only BSS, and activating both. All sparse symbols are randomly generated, with the random seeds for BSS varying across groups. @1, @2, and @3 correspond to BSS sparsity thresholds of 0.1, 0.3, and 0.5, respectively, while, within each group, the FC threshold increases incrementally (0.1, 0.2, 0.4, 0.6, 0.8). Our experiments show that, at the same sparsity level, FC achieves higher speedup than BSS. This is because FC requires only a single decoding operation on the CUDA cores, whereas BSS—despite having the same sparsity—requires multiple decoding operations along the reduction axis, which reduces efficiency. For the combined sparsity strategy, the speedup scales almost linearly with sparsity (approximately a 1:1 ratio). As an illustration, a combined sparsity of 90\% yields an observed speedup of about \(\sim9.4\times\) in Figure~\ref{exp:FlashOmni_attn_kernel}.

\subsection{\method Sparse GEMM-\(O\)}
\label{appendix_gemm_o}
\begin{figure}[t]
    \centering
    \includegraphics[width=\linewidth]{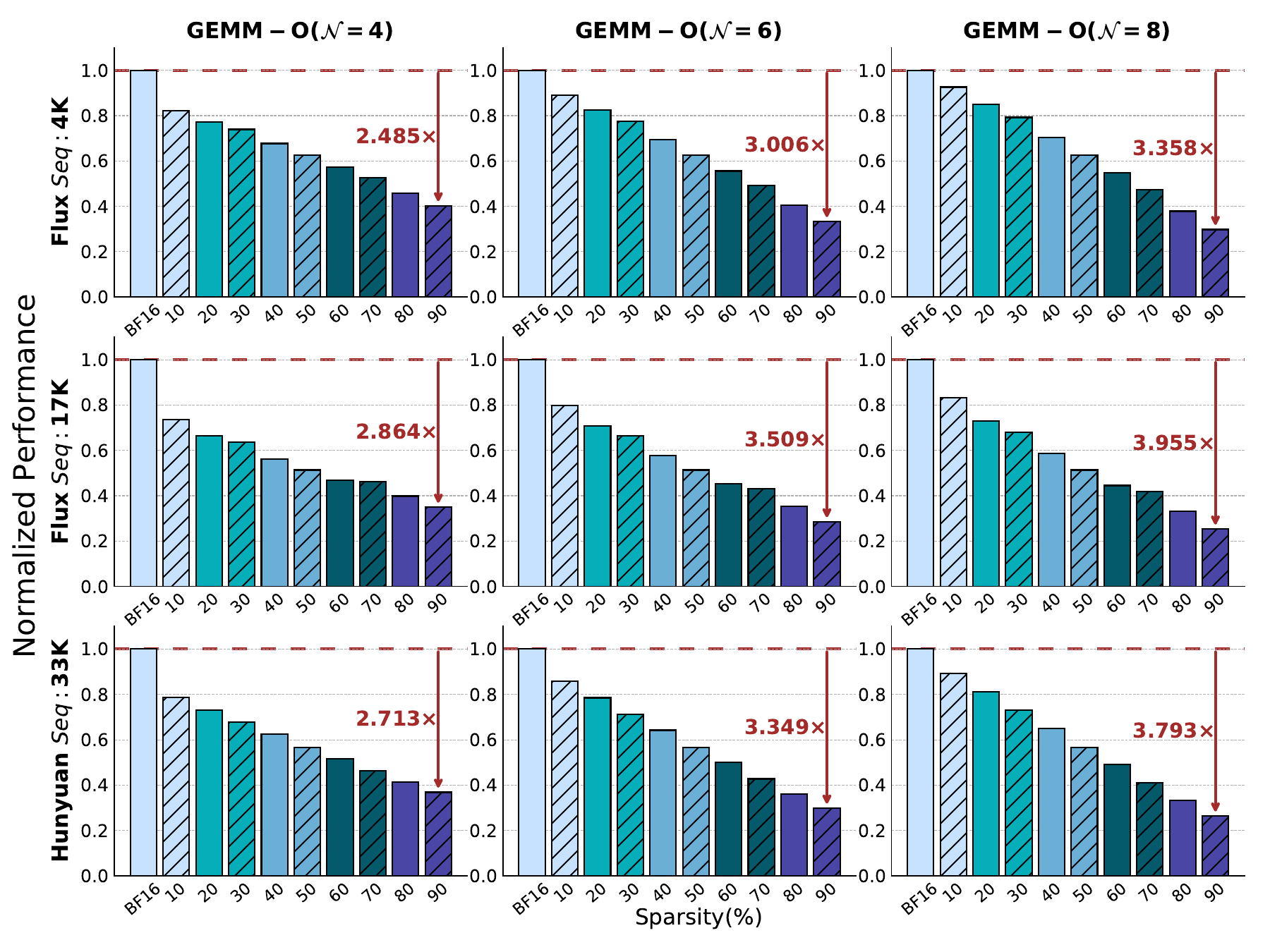}
    \vspace{-15pt}
	\caption{Normalized inference performance on NVIDIA A100 (BF16, measured by using cudaLib) at different sparsity levels for sparse GEMM-\(O\), whose  $\mathcal{N}$ includes 4,6,8.}
    \vspace{-10pt}
    \label{exp:FlashOmni_out_kernel}
\end{figure}

\looseness=-1
We further evaluated the GEMM-\(O\) speedup performance across three representative resolution settings for the generation tasks. As shown in the Figure~\ref{exp:FlashOmni_out_kernel}, for tasks at standard resolutions (e.g., 1K image generation in FLUX.1), the parallelism of the kernel is relatively limited, and the influence of CUDA Core decoding operations becomes higher. Consequently, the speedup is lower compared to the ultra-high-resolution scenarios. Nevertheless, the acceleration remains notable, achieving approximately \(2.5\times\)–\(3.4\times\) speedup across different \(\mathcal{N}\) settings. In the ultra-high-resolution cases, the speedup increases further, reaching avg \(2.7\times\)–\(3.9\times\).

\subsection{\method Programming Interface}

\definecolor{codeblue}{rgb}{0,0,0.5}
\definecolor{codeblue2}{rgb}{0,0,1}
\definecolor{lightgray}{rgb}{0.95, 0.95, 0.95}
\definecolor{mygreen}{rgb}{0.016, 0.44, 0.04}
\lstset{
  backgroundcolor=\color{lightgray},
  basicstyle=\fontsize{6.8pt}{6.8pt}\ttfamily\selectfont,
  columns=fixed,
  breaklines=true,
  captionpos=b,
  commentstyle=\fontsize{6.8pt}{6.8pt}\color{mygreen},
  keywordstyle=\fontsize{7.36pt}{7.36pt}\color{codeblue2},
    emph={flashomni},
    emphstyle={\bfseries\color[RGB]{59,93,128}},
}

\begin{lstlisting}[language=python]
import flashomni

# Attention Processor Modules
def __attn_wrapper__(self = AttnProcessor, task_info):
    self.attn_proc = flashomni.AttentionWrapper(task_info)

def __call__(self = AttnProcessor, attn, x, cache_dic):
  q = flashomni.to_q(cache_dic.sparse_symbols, x)                           # FlashOmni GEMM Q
  ...
  attn_out = self.attn_proc(q,k,v, cache_dic.sparse_symbols)                # FlashOmni Attention
  ...
  if cache_dic[type] == "update":
    # FlashOmni Sparse Symbols
    cache_dic.sparse_symbols = self.update_sparse_symbols(q, k)
    # Cache bias generation
    cached_bias = flashomni.to_out(attn_out, cache_dic.sparse_symbols)      # FlashOmni GEMM O
        
  out = flashomni.to_out(attn_out, cache_dic.sparse_symbols, cached_bias)   # FlashOmni GEMM O
  return out
\end{lstlisting}
\vspace{-1em} 

\newpage
\subsection{Supplementary Visualization Results}
\begin{figure}[h]
    \centering
    \includegraphics[width=\linewidth]{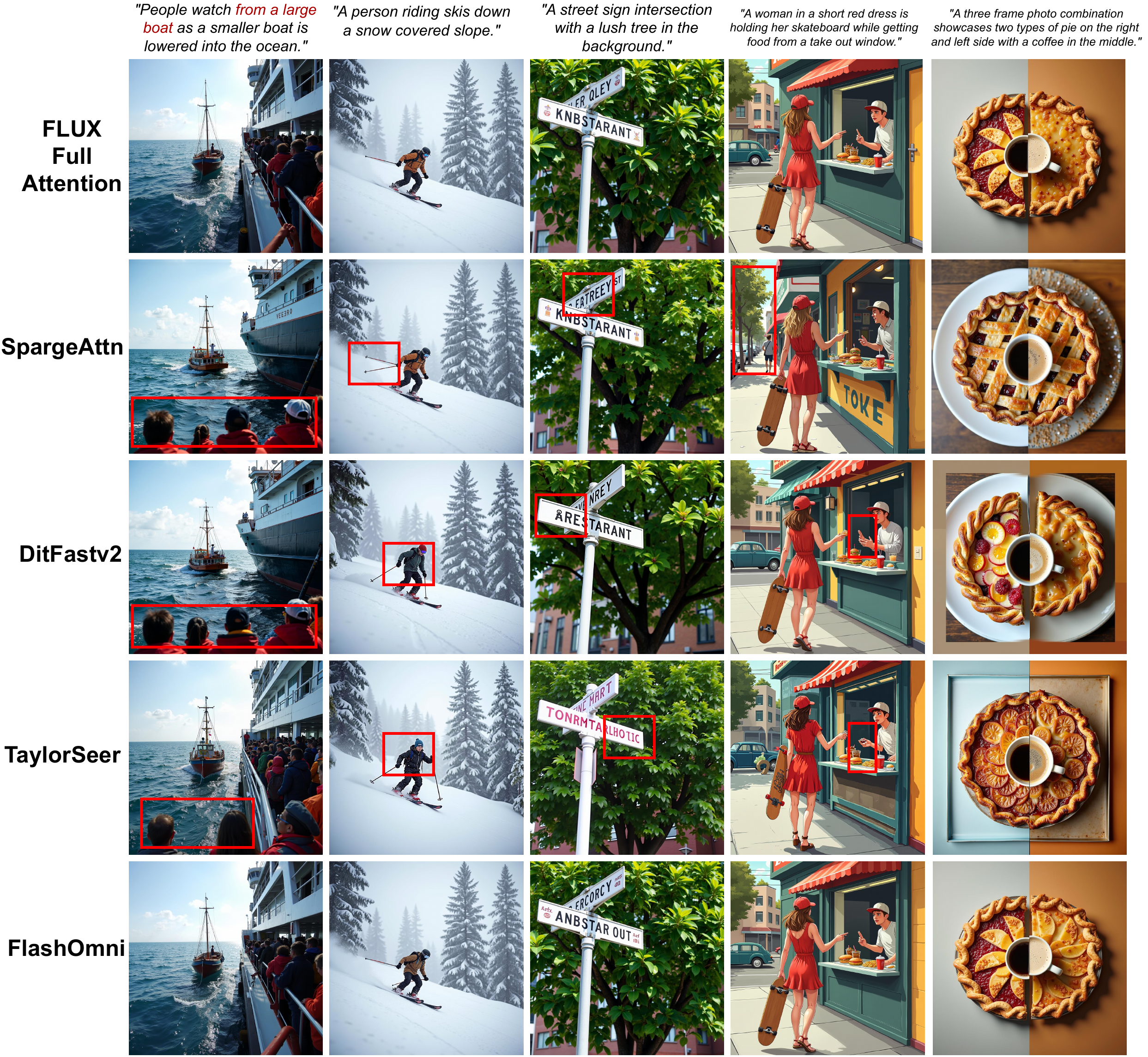}
    \vspace{-15pt}
	\caption{Visualization results for different acceleration methods on FLUX.1-dev.
    }
    \vspace{-10pt}
    \label{fig:fulu1}
\end{figure}
\begin{figure}[h]
    \centering
    \includegraphics[width=\linewidth]{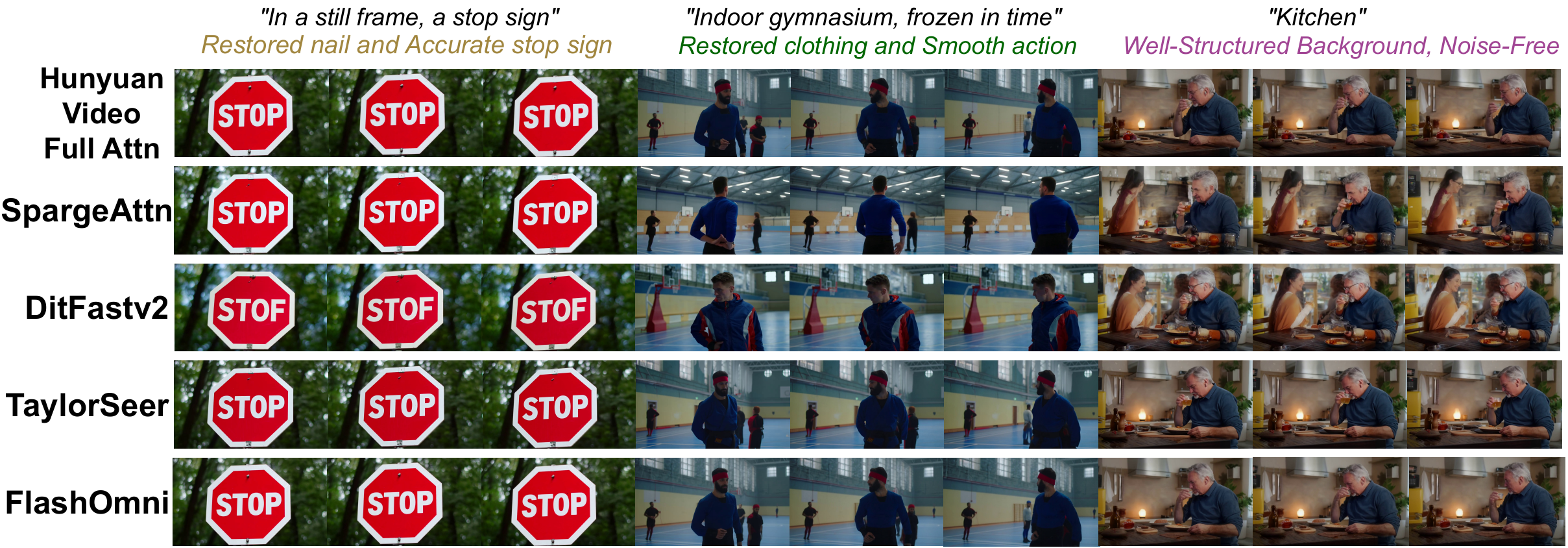}
    \vspace{-15pt}
	\caption{Visualization results for different acceleration methods on HunyuanVideo.
    }
    \vspace{-10pt}
    \label{fig:fulu2}
\end{figure}

\end{document}